\documentclass[letterpaper, 10 pt, conference]{ieeeconf}
\usepackage[dvipsnames]{xcolor}

\usepackage{graphicx}
\usepackage{subcaption}
\usepackage{cite}
\usepackage{picinpar}
\usepackage{url}
\usepackage{flushend}
\usepackage[utf8]{inputenc}
\usepackage{colortbl}
\usepackage{ulem}
\usepackage{multirow}
\usepackage{pifont}
\usepackage{color,soul}
\usepackage{alltt}
\usepackage[hidelinks]{hyperref}
\usepackage{enumerate}
\usepackage{siunitx}
\usepackage{breakurl}
\usepackage{epstopdf}
\usepackage{pbox}
\usepackage{lipsum}
\usepackage{colortbl}
\usepackage{booktabs}

\usepackage{textcomp}
\usepackage{tabularx}
\usepackage[export]{adjustbox}
\usepackage{stfloats} 
\usepackage{amsmath,amssymb,amsfonts}
\usepackage{array}
\usepackage{hyperref}
\usepackage{booktabs}
\usepackage[multiple]{footmisc}
\usepackage{amstext}
\usepackage{algpseudocode}
\usepackage{algorithm}
\usepackage{textcomp}
\usepackage{xcolor}

\newcolumntype{Y}{>{\centering\arraybackslash}X}
\IEEEoverridecommandlockouts
\begin{document}

\title{\LARGE \bf XIRVIO: Critic-guided Iterative Refinement for Visual-Inertial Odometry with Explainable Adaptive Weighting}

\author{Chit Yuen Lam$^{1}$, Ronald Clark$^{2}$ and Basaran Bahadir Kocer$^{3}$
\thanks{$^{1}$Chit Yuen Lam (\texttt{ian.lam@airbus.com}) is with Airbus, Pegasus House, Aerospace Avenue, Filton, Bristol, BS34 7PA. $^{2}$Authors are with the Department of Computer Science at the University of Oxford, UK. $^{3}$Authors are with the School of Civil, Aerospace and Design Engineering,
University of Bristol.}
}


\maketitle

\begin{abstract}
We introduce XIRVIO, a transformer-based Generative Adversarial Network (GAN) framework for monocular visual inertial odometry (VIO). By taking sequences of images and 6-DoF inertial measurements as inputs, XIRVIO's generator predicts pose trajectories through an iterative refinement process which are then evaluated by the critic to select the iteration with the optimised prediction. Additionally, the self-emergent adaptive sensor weighting reveals how XIRVIO attends to each sensory input based on contextual cues in the data, making it a promising approach for achieving explainability in safety-critical VIO applications. Evaluations on the KITTI dataset demonstrate that XIRVIO matches well-known state-of-the-art learning-based methods in terms of both translation and rotation errors. 
\end{abstract}


\section{Introduction}
Accurate and reliable state estimation is fundamental to the autonomy of robotic systems, but can be challenging when navigating cluttered indoor spaces, dynamic urban environments, and unstructured natural terrains like forests \cite{kocer2021forest,ho2022vision,lan2024aerial,bates2025leaf}. VIO leverages the complementary strengths of cameras and inertial measurement units (IMUs) to estimate the camera motion, but its performance is inherently tied to the reliability of each sensor under varying conditions. Visual data can be rich in features but is susceptible to challenges such as motion blur, occlusions, and low-light conditions. On the other hand, IMUs provide high-frequency motion updates but suffer from drift over time. The challenge lies in dynamically adapting the contribution of each sensor to ensure robust performance across diverse operational scenarios. 

\begin{figure}[t!]
    \centering
    \includegraphics[width=0.475\textwidth]{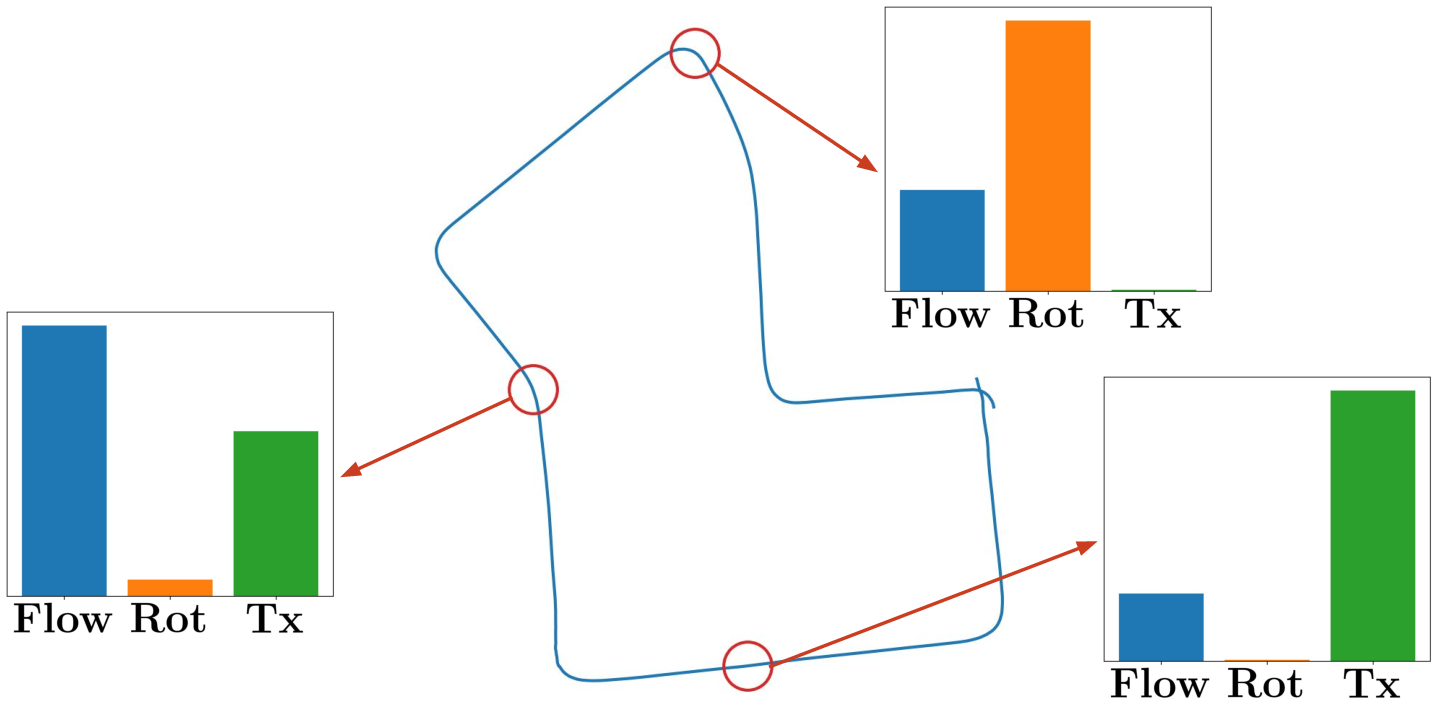}
    \caption{Overview of our self-emergent adaptive sensor weighting for a sample trajectory from the KITTI dataset. The stacked bars represent how the policy encoder dynamically allocates weights to each sensor modality. The labels "Flow", "Rot", and "Tx" on the bar charts denote optical flow, IMU rotation, and IMU translation respectively. The weights are normalised (min–max) within each modality for clarity.}
    \label{fig:overview_policy}
\end{figure}

In this paper, we introduce an adaptive sensor weighting approach that dynamically prioritises visual or inertial modalities based on environmental conditions as illustrated in Figure \ref{fig:overview_policy}. Central to our approach is a combination of traditional optimisation and modern learning-based techniques. Specifically, we introduce a Wasserstein GAN (WGAN) and use a WGAN loss together with pose regressive loss to train our model. The introduction of the WGAN critic not only allows us to improve the training, but also allows us to iteratively optimise the pose predictions at inference time. Effectively, the critic can be seen as a learned objective that overcomes the limitations of traditional objectives, such as photometric error, especially in textureless regions. This improves reliability, ensuring robust state estimation even in challenging conditions such as varying lighting and occlusions.


\section{Related Work}



VIO is a technique that combines visual data from cameras with inertial measurements from IMUs to estimate the camera's position and orientation over time \cite{scaramuzza2011visual,forster2014svo,qin2018vins}. This fusion leverages the strengths of both sensor types: cameras provide detailed environmental features, while IMUs offer high-frequency motion data \cite{leutenegger2015keyframe}. There are common geometric and learning-based approaches to solve the VIO problem. Geometric methods rely on traditional computer vision techniques, utilising feature extraction, epipolar geometry, and bundle adjustment to compute motion estimates \cite{orb,orbslam,sift,surf,klein2007parallel}. While learning-based methods mostly employ deep neural networks to infer motion directly from raw sensor data \cite{deepvo,clark2017vinet,learned_initial}. Despite the success of learning-based methods compared to geometric approaches, many existing learning-based approaches in the literature process data as streams without adequately evaluating degraded sensory inputs.

To handle imperfect sensory inputs in VIO, recent research on selective VIO emphasised the need to adaptively weigh or discard sensor inputs, visual and inertial, to cope with noisy environments, reduce computational overhead, and improve trajectory accuracy. For instance, \cite{carlone2018attention} proposed an anticipatory strategy that uses submodularity-driven feature selection to identify the most relevant visual cues under limited computational resources. With a similar motivation, \cite{liu2021atvio} introduced an attention-based VIO pipeline (ATVIO) that fuses camera and inertial data via a learnable attention block to emphasise useful sensory inputs while suppressing noisy or redundant signals. Other works focus on efficiency: \cite{yang2022efficient} described an adaptive deep-learning method that selectively disables the visual encoder when the IMU data alone suffices, saving up to 78\% of the computational load without significantly degrading odometry accuracy. In addition, \cite{chen2019selective} proposed a selective sensor fusion framework with “soft” and “hard” masking strategies to decide which modality, camera or IMU, contributes most to the current pose estimate by enhancing robustness to sensor failures and poor synchronisation. Although these methods demonstrate that selective processing of visual and inertial inputs can yield improved robustness and real-time performance, to the best of our knowledge there has been insufficient exploration of iterative refinement techniques for further improving pose estimations.

GAN have been explored in solving VIO problems. For example, GANVO learns to jointly predict depth map and pose via view reconstruction and without the aid of ground truth poses \cite{ganvo}. In addition, iterative refinement is a powerful strategy for improving pose estimation accuracy by revisiting initial predictions and incrementally reducing errors. For instance, DytanVO \cite{shen2023dytanvo} leveraged an iterative framework to jointly refine motion segmentation and ego-motion in scenes with extensive dynamic objects. Therefore, we propose a critic-guided iterative refinement approach to utilise the adversarial nature of GANs together with iterative refinement for VIO. Specifically, we introduce a generator trained with a hybrid loss that combines a mean square error (MSE) for pose loss with a WGAN generator loss, together with a critic trained with a WGAN critic loss weighted by the regressive pose loss from the generator. This enables XIRVIO to minimise the critic score during inference and thus converge to an optimised pose estimate. As a result, it can iteratively refine pose predictions even under ambiguity, hence delivering more reliable and robust VIO performance. 


\section{Methodology}

\begin{figure}[t!]
    \centering
    \includegraphics[width=0.475\textwidth]{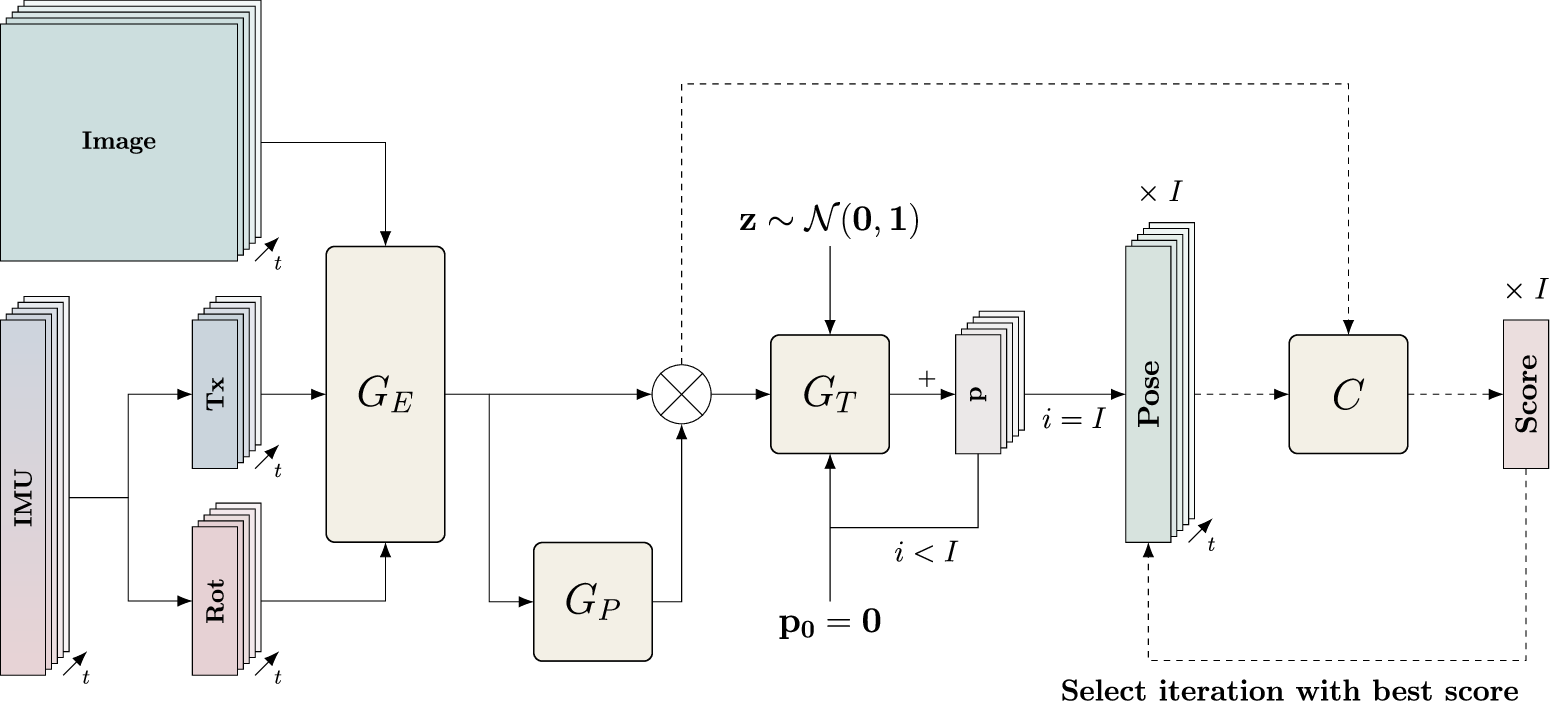}
    \caption{An overview of the XIRVIO architecture. The generator encodes the image and IMU inputs, produces an adaptive sensor weighting, and generates pose predictions iteratively. These poses are passed onto the critic together with the encoded vectors to obtain a critic score.}
    \label{fig:XIRVIO}
\end{figure}

Figure \ref{fig:XIRVIO} presents the overview of the XIRVIO architecture. XIRVIO is a conditional WGAN comprising a generator $G$ and a critic $C$. The generator is broken down into three main components - Feature Encoder $G_E$, Policy Encoder $G_P$, and the Generative-Iterative Pose Transformer $G_T$. The Feature Encoder first takes in the visual and inertial inputs and encodes them separately into sets of encoded features. These encoded features are passed onto the Policy Encoder to generate a weight for each sensory input and multiplies onto the encoded features. These weighted features are then concatenated and fed into the Generative-Iterative Pose Transformer, together with a random vector to generate and iteratively refine the pose estimations. All the pose iterations are fed into the critic together with the encoded features to obtain a critic score, where the iteration with the best critic score will be selected as the final pose estimation. This is illustrated in Figure \ref{fig:iterative_refinement} in detail.

\subsection{Generator}

\subsubsection{Input and Output Types}
The generator is the core of XIRVIO as it accepts sequences of images and 6-DoF IMU readings, encodes them into latent vectors, and outputs corresponding sequences of 6-DoF pose estimations. Besides, XIRVIO has the capability of accommodating various image formats like greyscale, depth maps, thermal maps, etc., through a simple casting layer. For the purposes of initial architectural exploration, this work focuses exclusively on RGB images. Nevertheless, this built-in flexibility opens doors for future enhancements, making the model versatile enough for broader application contexts, as long as it is adequately trained and fine-tuned.

\subsubsection{Feature Encoder $G_E$ -- Generating Conditions for WGAN}
The Feature Encoder of XIRVIO constitutes a flow-based visual encoder and an inertial encoder. The encoders take in image and IMU sequences respectively, then output a 1-dimensional latent vector of shape $[B, S, N_C]$, where $B$ is batch size, $S$ is sequence length, and $N_C$ is the vector size of one modality. Compared to other implementations of feature encoders, the inertial encoder encodes the translational and rotational components separately, allowing the Policy Encoder to dynamically weigh the translation and rotational modalities separately depending on the context of the sequence.
\paragraph{RAFT Optical Flow Visual Encoder}
In XIRVIO, optical flow is selected as the feature representing the visual modality. Using an iterative approach to refine its prediction, RAFT provides state-of-the-art optical flow predictions while maintaining a very small parameter footprint \cite{raft}. Since optical flow is not the major learning objective in the paper, a frozen RAFT-S model from PyTorch is utilised to provide accurate optical flow predictions out-of-the-box \cite{pytorch}. Moreover, the iterative nature of RAFT provides a flexible hyperparameter for the number of RAFT iterations, allowing fine tuning of computational cost to fulfil real-time computation needs. In this paper, we have selected the number of iterations to be 6 to balance between optical flow accuracy and computational time. As a final step, a ResNet backbone is then used to convert the latent RAFT outputs into 1-dimensional encoded vectors.
\paragraph{Inertial Encoder}
Compared to the visual encoder, the inertial component of the encoder is relatively straightforward. It is made up of several residual blocks and is notably smaller than the RAFT encoder. This is because IMU inputs are intrinsically simpler to encode than 2-dimensional visual data. The IMU encoder consists of two identical ResNet encoders to encode for translation and rotation separately. Moreover, separating the translational and rotational componentsallows the policy encoder to attend to these modalities dynamically.
\subsubsection{Policy Encoder $G_P$ -- Generating Self-emergent Adaptive Sensor Weighting} \label{sec:policy}
The Policy Encoder is the key area where the model autonomously learns to weigh different modalities depending on the context of the sequence. It is worth noting that the weights are not only learned autonomously without any guides in the loss function, but these self-learned weights are also human-interpretable at the same time. The Policy Encoder collects the concatenated latent vectors of the modalities, and passes it through a ResNet backbone to produce a weight vector of dimensions $[B, C, S]$, where $C$ is the number of modalities. These weights will be passed through a Softplus activation to convert the logits into unbounded positive weights. It is worth noting that the Softplus activation used in this paper has been slightly modified to parametrically shift the intercept to 1 regardless of the hyperparameter of the activation's softness.

These individual weights will be multiplied onto their respective modalities, forming a gated-attention mechanism and producing weighted feature vectors before passing them onto the pose estimator. This not only allows the features of different modalities to be weighted dynamically but also makes the model predictions more explainable as policy weightings are more intuitive to understand.

\subsubsection{Generative-Iterative Pose Transformer $G_T$} \label{sec:informer}
The Generative-Iterative Pose Transformer is the area where the generative element of GANs is introduced. It comprises a transformer architecture that takes in a concatenated time series of weighted features, random vectors, and a pose accumulator.
\paragraph{Iterative Refinement of Pose Estimation}
The Generative-Iterative Pose Transformer uses an iterative refinement method to improve the prediction of the pose iteratively. This is achieved by having a global variable $p$, known as the pose accumulator, to sum up the predictions for $I$ iterations. For each iteration, the Pose Transformer does not predict the pose directly, but instead uses a "delta pose" $\Delta p_i$ to refine the pose accumulator as shown in equation \eqref{eq:pose_accumulation}. Algorithm \ref{alg:deltapose} also describes how the Generative-Iterative Pose Transformer can iteratively refine the pose.
\begin{align} \label{eq:pose_accumulation}
    {\rm pose} &= \sum_{i=0}^{I} \Delta p_{i}
\end{align}
While Algorithm \ref{alg:deltapose} demonstrates the capability of iterative refinement, there is no guarantee that the final iteration is the best pose, like all other optimisation methods. As a result, a critic can be used for the self-evaluation of the  estimation to obtain the pose with the best critic score. This will be further discussed in Section \ref{sec:critic}.

\begin{algorithm}[t!]
    \caption{Iterative refinement and self-evaluation of the generative-iterative pose transformer} \label{alg:deltapose}
    \begin{algorithmic}
    \State $\mathbf{p} \gets \mathbf{0}$  \Comment{Initialise global pose accumulator with zeros}
    \State $p_i \gets 0$  \Comment{Initialise pose accumulator with zeros}
    \State $\Delta p_i \gets 0$  \Comment{Initialise "delta pose" with 0}
    \State $\tilde{x} \gets G_E(x) \cdot G_P(G_E(x))$  \Comment{Encode model inputs}
    \For{$i < I$}
        \State $\Delta p_i \gets G_T(z, \tilde{x}, p)$
        \State $p_i \gets p_i + \Delta p_i $
        \State $\mathbf{p}_i \gets p_i$ \Comment{Store the pose of each iteration}
    \EndFor
    
    \end{algorithmic}
\end{algorithm}

XIRVIO uses the Informer as a blueprint for the Generative Pose Transformer given its excellence in generating time-series predictions \cite{informer}. However, its encoder and decoder underwent significant modifications, especially regarding LayerNorm placement. Informer uses the vanilla Post-LN architecture \cite{vaswani}, where LayerNorm is located after the residual branch joining \cite{informer}. When training XIRVIO with this arrangement, gradient and convergence issues were observed, preventing the model from being trained reliably across multiple hyperparameter configurations. This echoes the known Post-LN challenges discussed in \cite{xformer_difficulties}, and as a result, the layers have been converted to a Pre-LN arrangement as described by \cite{preln}. Although this Pre-LN arrangement improves model convergence, the model capacity might be limited in deeper configurations, though the configuration used by XIRVIO is not necessarily deep \cite{xformer_difficulties}. In addition, a conditioning layer is added to the transformer to provide conditioning vectors to the outputs of each attention layer. This idea migrated from \textit{et al.} to provide context to visual inputs \cite{film}. Contrary to this work which uses words as the visual context for a static image, XIRVIO uses the input IMU sequence to provide context to the output sequences of the individual attention layers, before being projected to the 6DoF pose estimation.

\begin{figure}[b!]
    \centering
    \includegraphics[width=0.475\textwidth]{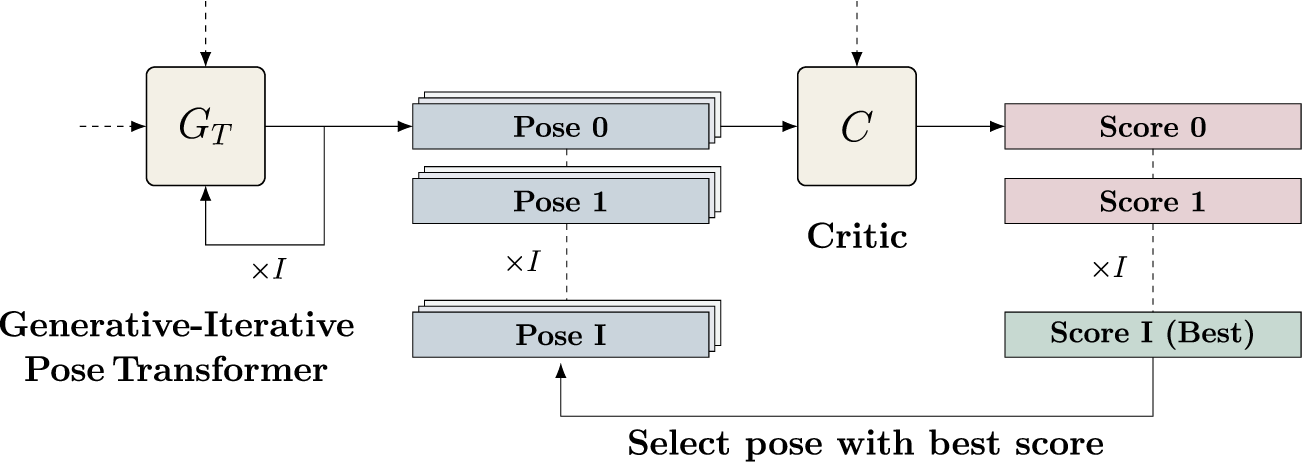}
    \caption{A simplified overview of critic-guided iterative refinement. The Generative-Iterative Pose Transformer $G_T$, generates and iteratively refines the pose estimations. All the pose iterations are evaluated by the critic $C$ to obtain a critic score, and the iteration with the best critic score will be selected as the final pose estimation.}
    \label{fig:iterative_refinement}
\end{figure}

\subsection{Critic} \label{sec:critic}
The architecture of critic is a simplified version of the generator. It consists the same encoder architecture as $G_T$ to predict the realness of each pose sequence, given the input pose sequence together with the encoded features from the generator. As XIRVIO follows a WGAN-GP training scheme, the output of the critic is unbounded \cite{wgangp}. Using the critic, the network can self-evaluate the quality of the generated poses of each iteration, and consequently select the best pose from the array of pose estimations from each iteration.
\paragraph{Self-evaluation of Iterative Pose Refinement}
In the Generative-Iterative Pose Transformer, while it iteratively refines the pose prediction, there is no guarantee that the pose estimation is the most converged at the final iteration, but probably the intermediate iterations instead. Therefore, not only can the critic be used to help train the generator for a more realistic pose, but also it can be used to select the best iteration for pose estimation. Figure \ref{fig:iterative_refinement} provides a simplified graphical illustration on how the iterations of pose estimation are evaluated by the critic to select the iteration with the optimal critic score. Algorithm \ref{alg:selfeval} also demonstrates a basic logic between the iterative refinement of the Pose Transformer and the self-evaluation of the critic.
\begin{algorithm}[t!]
    \caption{Self-evaluation of the generative-iterative pose transformer} \label{alg:selfeval}
    \begin{algorithmic}
    \State $\mathbf{p} \gets \mathbf{0}$  \Comment{Initialise global pose accumulator with zeros}
    \State $\mathbf{c} \gets \mathbf{0}$  \Comment{Initialise global critic score with zeros}
    \State $p_i \gets 0$  \Comment{Initialise pose accumulator with zeros}
    \State $\tilde{x} \gets G_E(x) \cdot G_P(G_E(x))$  \Comment{Encode model inputs}
    \For{$i < I$}
        \State $p_i \gets p_i + G_T(z, \tilde{x}, p_i) $
        \State $\mathbf{p}_i \gets p_i$ \Comment{Store the pose of each iteration}
    \EndFor
    \State $\mathbf{c} \gets C(x, \mathbf{p})$  \Comment{Compute critic score for all pose iterations}
    \State $j \gets \textrm{arg}\,\textrm{max}(\mathbf{c})$ \Comment{Select pose with best critic score}
    \State \Return $\mathbf{p}_{j}$
    \end{algorithmic}
\end{algorithm}

\subsection{Training}
The XIRVIO training scheme is similar to the typical WGAN-GP training, where the generator and the critic are trained for $m$ and $n$ steps per batch \cite{wgangp}. However, the generator objective is modified to include the iterative and regressive nature of pose estimation. In addition, the Wasserstein Difference of the critic is also weighted according to the mean squared error of the pose estimation during the training process. This is to provide context to the critic regarding the "realness" of the pose input.

\subsubsection{Loss Function}
In the following sections, the training objectives of the generator and critic are described. We first establish the following equations containing the common terms seen in Equations \eqref{eq:pose_t}, \eqref{eq:pose_r}, \eqref{eq:gen_critic}, \eqref{eq:critloss}, and \eqref{eq:pose_weight}:
\begin{align}
    \mathbf{\tilde{x}} &= G_E(\mathbf{x}) \label{eq:encode}\\
    \mathbf{\hat{y}} &= G(\mathbf{z}, \mathbf{x}),\;\;\;\mathbf{z} \sim \mathcal{N}(\mathbf{0}, \mathbf{1}) \label{eq:genpass}
\end{align}
In Equations \eqref{eq:encode} and \eqref{eq:genpass}, $\mathbf{x}$ refers to the generator inputs, which are images and IMUs in XIRVIO; $\mathbf{\tilde{x}}$ denotes the latent vectors encoded by the Feature Encoder $G_E$; $\mathbf{\hat{y}}$ denotes the pose estimation synthesised by the generator $G$. It is worth noting that $\mathbf{y}$ refers to the ground truth pose.
\paragraph{Generator}
The loss function of the generator consists of three elements: pose loss $\mathcal{L}_{pose}$, and critic loss $\mathcal{L}_C$ as shown in Equation \eqref{eq:genloss}. $\mathcal{L}_{pose}$ is further scaled by a constant factor of $\alpha$ to balance between the magnitudes of critic loss and the regressive loss.
\begin{align}
    \mathcal{L}_G &= \alpha \mathcal{L}_{pose} + \mathcal{L}_{critic} \label{eq:genloss}
\end{align}
$\mathcal{L}_{pose}$ is also further divided into translational and rotational components $\mathcal{L}_{p_t}$ and $\mathcal{L}_{p_r}$ respectively in equation \eqref{eq:pose_t_r}, where $\mathcal{L}_{p_r}$ is scaled by a constant factor $\beta$ to emphasise on rotational errors.
\begin{align}
    \mathcal{L}_{pose} &= \mathcal{L}_{p_t} + \beta \mathcal{L}_{p_r} \label{eq:pose_t_r}
\end{align}
In order to train the generator to iteratively refine its prediction, $\mathcal{L}_{pose}$ and $\mathcal{L}_{critic}$ are weighted in the "iteration" dimension with $\mathbf{\lambda}$ as defined in equation \eqref{eq:lambda}. (i.e. given an arbitrary shape $[B, I, ...]$, where $B$ and $I$ are batch size and iteration dimension, respectively, the losses will be weighted in the $I$ dimension). 
\begin{align}
    \boldsymbol{\lambda} &= \frac{1}{\int_0^I \gamma^i \;di} \big[ \gamma^{I}, \gamma^{I-1}, ..., \gamma^{0} \big], \;\;\textrm{where}\; 0<\gamma<1 \label{eq:lambda} \\
    \mathcal{L}_{p_t} &= \underset{\mathbf{z} \sim \mathcal{N}(\mathbf{0}, \mathbf{1})}{\mathbb{E}} \Big[ \boldsymbol{\lambda} \big( \mathbf{y}_t - \mathbf{\hat{y}}_t \big)^2 \Big] \label{eq:pose_t}\\
    \mathcal{L}_{p_r} &= \underset{\mathbf{z} \sim \mathcal{N}(\mathbf{0}, \mathbf{1})}{\mathbb{E}} \Big[ \boldsymbol{\lambda} \big( \mathbf{y}_r - \mathbf{\hat{y}}_r \big)^2 \Big] \label{eq:pose_r}\\
    \mathcal{L}_{critic} &= \underset{\mathbf{z} \sim \mathcal{N}(\mathbf{0}, \mathbf{1})}{\mathbb{E}} \Big[ \boldsymbol{\lambda} \, C(\mathbf{\tilde{x}}, \mathbf{\hat{y}}) \Big] \label{eq:gen_critic} \\
\end{align}
In this paper, the constants $\alpha$, $\beta$, $\gamma$, and $I$ are set to $100$, $10$, and $0.8$, and $16$ respectively.graph{Critic}
The loss function of our critic as shown in Equation \eqref{eq:critloss} is similar to that of a vanilla WGAN-GP critic \cite{wgangp} where $\psi = 10$. However, in addition to the vanilla WGAN-GP critic loss, a regressive term $\mathbf{w}$ obtained from the MSE of real and generated poses is applied to the critic score of the generated pose as shown in Equation \eqref{eq:pose_weight}.
\begin{align}
    \mathcal{L}_C &= \mathbf{w}C(\mathbf{\hat{y}}) - C(\mathbf{y}) + \psi L_{GP} \label{eq:critloss}\\
    \mathcal{L}_{GP} &= \underset{\mathbf{z} \sim \mathcal{N}(\mathbf{0}, \mathbf{1})}{\mathbb{E}} \Big[ \big( || \nabla_{\!\mathbf{\hat{y}}} \,C(\mathbf{\hat{y}}) ||_2 - 1 \big)^2 \Big] \\
    \mathbf{w} &= (\mathbf{y} - \mathbf{\hat{y}})^2 \label{eq:pose_weight}
\end{align}
This is possible as pose estimation from a given sequence of images and IMUs is essentially a regression task. With this regressive term, the critic gains knowledge on how "fake" a generated sample really is and optimise accordingly. Moreover, this enables the critic to be used for self-evaluation of the generator which optimises the pose iteratively and pick the best iteration accordingly. It has been observed empirically that without this term, critic will emphasise its training on the iterations with a high regressive loss (i.e. Iteration 0), making the early iterations having a low critic loss and later iterations having a high critic loss. In the training of critic, ll iterations will be used as a training sample (i.e. there will be $B \times I$ samples for a generator output of $[B, I, ...]$)


\section{Results}

\subsection{Configuration}

\newcommand{\cpu}{AMD Ryzen 9 7950X}
\newcommand{\ram}{64GB DDR5 6000Mb/s}
\newcommand{\gpu}{NVIDIA RTX 4090}
\newcommand{\ssd}{WD Black SN850X 4TB}

\newcommand{\modelsizes}{S/M/L}
\newcommand{\imgsize}{(256, 512)}
\newcommand{\seqlen}{4}
\newcommand{\vectorsize}{256/384/512}
\newcommand{\nconds}{4/6/8}
\newcommand{\ngen}{4}
\newcommand{\ndis}{4}
\newcommand{\ngdm}{512/768/1024}
\newcommand{\nddm}{512/768/1024}
\newcommand{\flow}{RAFT-S \cite{raft} \cite{pytorch} \cite{torchvision}}

\newcommand{\seed}{0}
\newcommand{\dropout}{0.1}
\newcommand{\aug}{None}
\newcommand{\batchsize}{64}
\newcommand{\opt}{AdamW \cite{adamw} \cite{pytorch}}
\newcommand{\lr}{0.0001}
\newcommand{\decay}{0.001}
\newcommand{\betas}{(0.5, 0.9) \cite{wgangp}}
\newcommand{\grepeat}{1}
\newcommand{\drepeat}{2}
\newcommand{\scheduler}{ReduceLROnPlateau \cite{pytorch}}
\newcommand{\patience}{10}
\newcommand{\schedfactor}{0.5}
\newcommand{\kittitrain}{00-02,06,08,09}
\newcommand{\kittieval}{05,07,10}

Table \ref{tab:setup} details the hardware configuration, model architecture, and training configuration used for training and evaluation. Unless otherwise stated, all experiments in subsequent sections adhere to this setup. It is worth mentioning that multiple model variants of different vector widths and layer depths, denoted by the suffixes "-S/M/L", have been trained to evaluate how the model performs under lightweight configurations. Due to the stochastic nature of GANs and the non-determinism of the kernels used in XIRVIO without deterministic alternatives, a Monte-Carlo approach is taken and the results are obtained by evaluating the sequences for 10 repeats. Unless otherwise specified, all XIRVIO variants are trained and evaluated with 16 iterations. The number of iterations used will be denoted as a subscript at the end of the model name (e.g. $\textrm{XIRVIO-L}_{16}$ denotes XIRVIO Large variant trained with 16 iterations).

\begin{table}[t!]
    \centering
    \caption{Key setup for hardware, model architecture, and model training.}
    \label{tab:setup}
    \scalebox{0.675}{
    \centering
    \renewcommand{\arraystretch}{1.1}
    \begin{tabular}{ll|ll}
        \hline \hline
        \multicolumn{4}{c}{\textbf{System Configuration}} \\
        \hline
        
        {CPU} & \cpu & {GPU} & \gpu \\
        {Memory} & \ram & {Storage} & \ssd \\

        \hline
        \multicolumn{4}{c}{\textbf{Model Architecture}} \\
        \hline

        {Model Variant} & \modelsizes & {Image Size} & \imgsize \\
        {Vector Width} & \vectorsize & {Sequence Length} & \seqlen \\
        {Cond. Encoder Layers} & \nconds & {Optical Flow} & \flow \\
        {Generator Layers} & \ngen & {Critic Layers} & \ndis \\
        {Gen. Hidden Width} & \ngdm & {Critic Hidden Width} & \nddm \\

        \hline
        \multicolumn{4}{c}{\textbf{Training Configuration}} \\
        \hline

        Batch Size & \batchsize & Random Seed & \seed \\
        Dropout & \dropout & Image Augmentation & \aug \\
        {Optimiser} & \opt & {LR Scheduler} & \scheduler \\
        {Betas} & \betas & {Scheduler Patience} & \patience \\
        {Weight Decay} & \decay & {Scheduler Factor} & \schedfactor \\
        {Generator LR} & \lr & {Generator Steps / Batch} & \grepeat \\
        {Critic LR} & \lr & {Critic Steps / Batch} & \drepeat \\
        {KITTI Train \cite{kittiodometry}} & \kittitrain & {KITTI Eval. \cite{kittiodometry}} & \kittieval \\
        
        \hline \hline
    \end{tabular}}
\end{table}

\subsubsection{KITTI Dataset} In terms of KITTI sequences for training and evaluation, sequences 00-02, 06, 08, and 09 are used for training, while sequences 05, 07, and 10 are reserved for evaluation. This provides an identical set of evaluation sequences as VSVIO, VIFT, and ATVIO \cite{vsvio} \cite{vift} \cite{atvio}. However, it is worth noting that sequence 04 is further removed from the training dataset, as sequence 04 is a straight-line sequence and has a drastically different distribution in the ground truth pose as well as the IMU inputs. Not only does this removal improve training speed, but the distributions of the training and evaluation datasets are also more aligned as a result.

\subsubsection{Metrics} In the evaluation of model prediction performance, four metrics, namely relative translation error ($t_{rel}$), relative rotation error ($t_{rel}$), translational RMSE ($t_{rmse}$), and rotational RMSE ($r_{rmse}$), will be used. However, only the relative errors will be the focus of performance evaluation in this section as RMSE is not as representative in terms of the overall trajectory deviation.

\subsection{Performance}







\newcommand{\largefivetrel}{1.77}
\newcommand{\largefiverrel}{0.62}

\newcommand{\largeseventrel}{1.53}
\newcommand{\largesevenrrel}{0.88}

\newcommand{\largetentrel}{3.70}
\newcommand{\largetenrrel}{1.22}

\newcommand{\mediumfivetrel}{2.24}
\newcommand{\mediumfiverrel}{0.97}

\newcommand{\mediumseventrel}{2.12}
\newcommand{\mediumsevenrrel}{0.90}

\newcommand{\mediumtentrel}{5.44}
\newcommand{\mediumtenrrel}{0.98}

\newcommand{\smallfivetrel}{2.42}
\newcommand{\smallfiverrel}{0.84}

\newcommand{\smallseventrel}{2.65}
\newcommand{\smallsevenrrel}{1.10}

\newcommand{\smalltentrel}{6.34}
\newcommand{\smalltenrrel}{1.24}

\newcommand{\vsviofivetrel}{2.01}
\newcommand{\vsviofiverrel}{0.75}

\newcommand{\vsvioseventrel}{1.79}
\newcommand{\vsviosevenrrel}{0.76}

\newcommand{\vsviotentrel}{3.41}
\newcommand{\vsviotenrrel}{1.08}

\newcommand{\dviofivetrel}{2.86$^{*}$}
\newcommand{\dviofiverrel}{2.32$^{*}$}

\newcommand{\dvioseventrel}{2.71$^{*}$}
\newcommand{\dviosevenrrel}{1.66$^{*}$}

\newcommand{\dviotentrel}{0.85}
\newcommand{\dviotenrrel}{1.03}

\newcommand{\sviofivetrel}{0.89$^{*}$}
\newcommand{\sviofiverrel}{0.63$^{*}$}

\newcommand{\svioseventrel}{0.91$^{*}$}
\newcommand{\sviosevenrrel}{0.49$^{*}$}

\newcommand{\sviotentrel}{1.81}
\newcommand{\sviotenrrel}{1.30}

\newcommand{\atviofivetrel}{4.93}
\newcommand{\atviofiverrel}{2.40}

\newcommand{\atvioseventrel}{3.78}
\newcommand{\atviosevenrrel}{2.59}

\newcommand{\atviotentrel}{5.71}
\newcommand{\atviotenrrel}{2.96}

\newcommand{\viftfivetrel}{2.02}
\newcommand{\viftfiverrel}{0.53}

\newcommand{\viftseventrel}{1.75}
\newcommand{\viftsevenrrel}{0.47}

\newcommand{\vifttentrel}{2.11}
\newcommand{\vifttenrrel}{0.39}

\newcommand{\wganvofivetrel}{7.01}
\newcommand{\wganvofiverrel}{3.85}

\newcommand{\wganvoseventrel}{7.71}
\newcommand{\wganvosevenrrel}{3.79}

\newcommand{\wganvotentrel}{10.89}
\newcommand{\wganvotenrrel}{5.03}

\newcommand{\best}[1]{\textcolor{blue}{\textbf{#1}}}

Table \ref{tab:perf} demonstrates the performance comparison of our XIRVIO against other learning-based methods for VIO using KITTI dataset sequences 05, 07, and 10. Although sequences 05, 07, and 10 align with the evaluation dataset of methods such as VSVIO, VIFT, and ATVIO, it is worth noting that some methods like DeepVIO and SelfVIO use sequence 00-08 as the training dataset \cite{deepvio} \cite{selfvio}. This is also marked accordingly in the table.

\begin{table}[t!]
    \renewcommand{\arraystretch}{1.0}
    \centering
    \caption{Comparison of KITTI benchmarks against other learning-based methods for VIO. The bold text denotes the best score of that sequence (excluding methods that use the sequence as the training dataset).}
    \label{tab:perf}
    \scalebox{0.8}{
    \begin{tabular}{l|ll|ll|ll}
        \hline \hline
        \multirow{2}{*}{\textbf{Method}} & \multicolumn{2}{c|}{\textbf{05}} & \multicolumn{2}{c|}{\textbf{07}} & \multicolumn{2}{c}{\textbf{10}} \\
        \cline{2-7}
         & $t_{rel} (\%)$ & $r_{rel} (^\circ)$ & $t_{rel} (\%)$ & $r_{rel} (^\circ)$ & $t_{rel} (\%)$ & $r_{rel} (^\circ)$ \\
        \hline
        VSVIO \cite{vsvio} & \vsviofivetrel & \vsviofiverrel & \vsvioseventrel & \vsviosevenrrel & \vsviotentrel & \vsviotenrrel \\
        VIFT \cite{vift} & \viftfivetrel & \best{\viftfiverrel} & \viftseventrel & \best{\viftsevenrrel} & \vifttentrel & \best{\vifttenrrel} \\
        ATVIO \cite{atvio} & \atviofivetrel & \atviofiverrel & \atvioseventrel & \atviosevenrrel & \atviotentrel & \atviotenrrel \\
        SelfVIO \cite{selfvio} & \sviofivetrel & \sviofiverrel & \svioseventrel & \sviosevenrrel & \sviotentrel & \sviotenrrel \\
        DeepVIO \cite{deepvio} & \dviofivetrel & \dviofiverrel & \dvioseventrel & \dviosevenrrel & \best{\dviotentrel} & \dviotenrrel \\
        WGANVO \cite{wganvo} & \wganvofivetrel & \wganvofiverrel & \wganvoseventrel & \wganvosevenrrel & \wganvotentrel & \wganvotenrrel \\
        \hline
        $\textrm{XIRVIO-S}_{16}$ & \smallfivetrel & \smallfiverrel & \smallseventrel & \smallsevenrrel & \smalltentrel & \smalltenrrel \\
        $\textrm{XIRVIO-M}_{16}$ & \mediumfivetrel & \mediumfiverrel & \mediumseventrel & \mediumsevenrrel & \mediumtentrel & \mediumtenrrel \\
        $\textrm{XIRVIO-L}_{16}$ & \best{\largefivetrel} & \largefiverrel & \best{\largeseventrel} & \largesevenrrel & \largetentrel & \largetenrrel \\
        \hline \hline
        \multicolumn{7}{l}{$^{*}$ sequence is part of the method's training dataset}
    \end{tabular}}
\end{table}

From Table \ref{tab:perf}, it can be seen that our method has matched, if not exceeded, the performance of some other state-of-the-art learning-based VIO methods across multiple variants of XIRVIO. This is encouraging as it shows our method exhibits capabilities of self-optimisation during inference and generating self-emergent policies, apart from just providing state-of-the-art performance. 

\subsection{Critic-guided Iterative Refinement}

\begin{figure}[t!]
    \centering
    \begin{subfigure}[t]{0.42\textwidth}
        \centering
        \includegraphics[width=\textwidth]{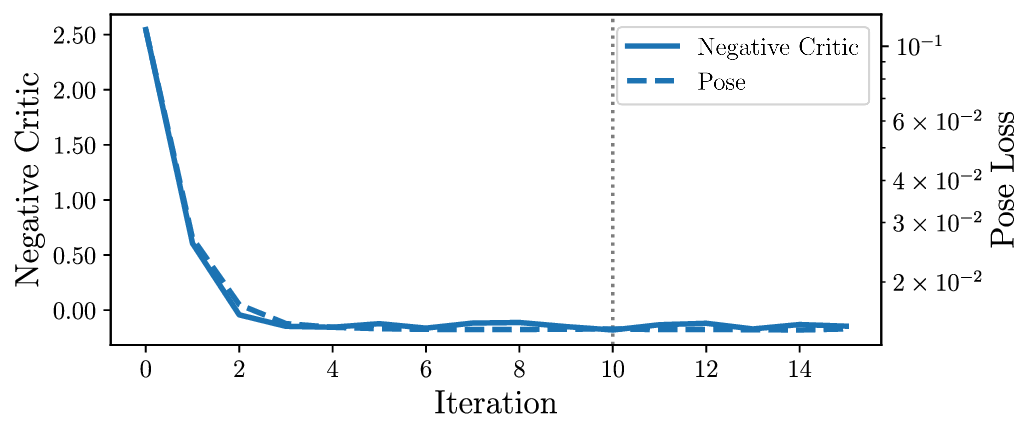}
        \caption{Training.}
        \label{fig:loss_per_iter_train}
    \end{subfigure}

    \begin{subfigure}[t]{0.42\textwidth}
        \centering
        \includegraphics[width=\textwidth]{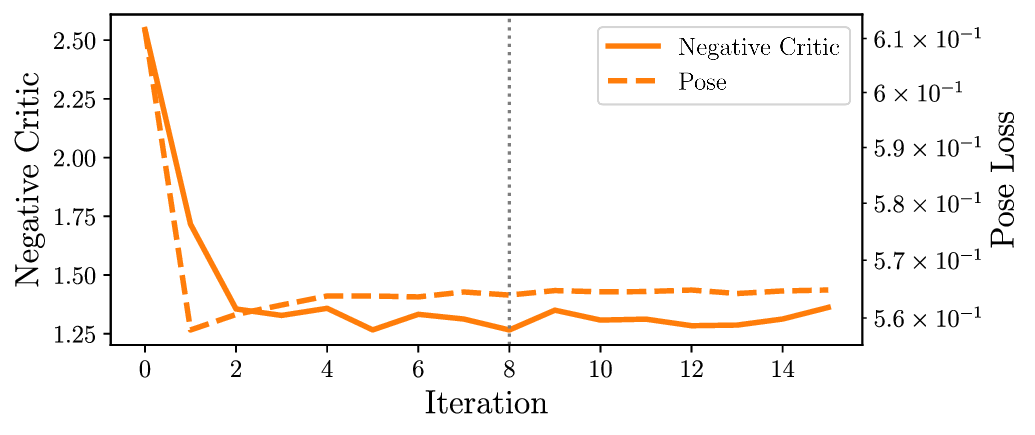}
        \caption{Evaluation.}
        \label{fig:loss_per_iter_eval}
    \end{subfigure}
    \caption{Variation of pose loss and negative critic score per iteration.}
    \label{fig:loss_per_iter}
\end{figure}

Another key novelty of the XIRVIO is the Critic-guided Iterative Refinement. This involves the generator iteratively predicting the pose for $I$ iterations, and the critic evaluates all of these iterations and then selects the generator iteration with the best critic score. This section investigates how iterative refinement impacts the model performance. Figure \ref{fig:loss_per_iter} demonstrates the variation pose loss and negative critic score with the number of iterations (Note: negative critic score is used here as a more positive critic score indicates a higher quality sample as it follows a typical WGAN-GP training \cite{wgangp}). Across both training and evaluation, both pose loss and negative critic scores converged after about 5 iterations, and the negative critic score exhibited a larger fluctuation than the pose loss. Moreover, during evaluation, the pose loss reached a minimum at $I=1$ and before increasing slightly and settling, while the critic loss maintains the typical convergence behaviour. 

\begin{figure}[b!]
    \centering
    \begin{subfigure}[t]{0.48\textwidth}
        \centering
        \includegraphics[width=\textwidth]{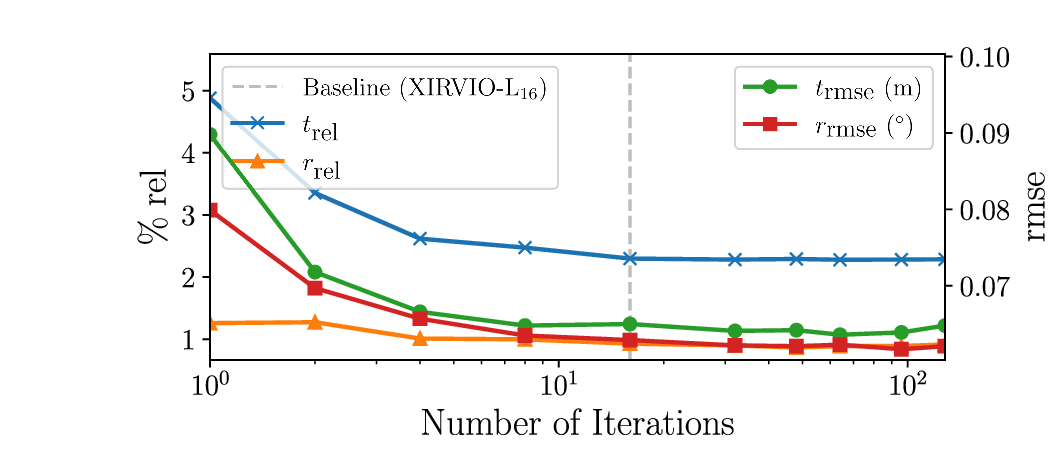}
        \caption{KITTI evaluation metrics \cite{kittiodometry}.}
        \label{fig:loss_vs_iter}
    \end{subfigure}

    \begin{subfigure}[t]{0.48\textwidth}
        \centering
        \includegraphics[width=\textwidth]{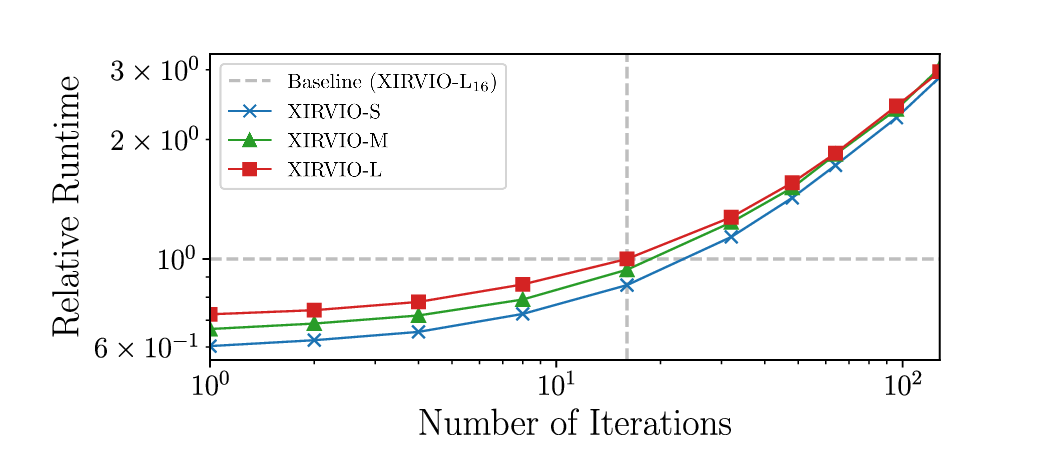}
        \caption{Model runtime relative to baseline XIRVIO-L with 16 iterations.}
        \label{fig:runtime_vs_iter}
    \end{subfigure}
    \caption{Variation of model performance against the number of iterations.}
    \label{fig:metrics_vs_iter}
\end{figure}

As part of the ablation studies for iterative refinement, the number of iterations used during evaluation has been investigated with a reference model ($\textrm{XIRVIO-L}_{16}$). Figure \ref{fig:loss_vs_iter} shows the variation of $t_{rel}$, $r_{rel}$, $t_{rmse}$, and $r_{rmse}$ with the log number of iterations ($\log_2i$). It can be observed that the prediction performance improves with the number of iterations. However, it can be observed that the improvement of model metrics comes with a price of computational cost as shown in Figure \ref{fig:runtime_vs_iter}, where the relative runtime increases exponentially with the number of iterations. Therefore, during deployment of XIRVIO the number of maximum iterations must be selected carefully to balance between the prediction accuracy and the need for low-latency pose prediction.

\subsection{Self-emergent Explainable Policy}

\begin{figure}[t!]
    \centering
    
    \begin{subfigure}[t]{0.225\textwidth}
        \centering
        \includegraphics[width=\textwidth]{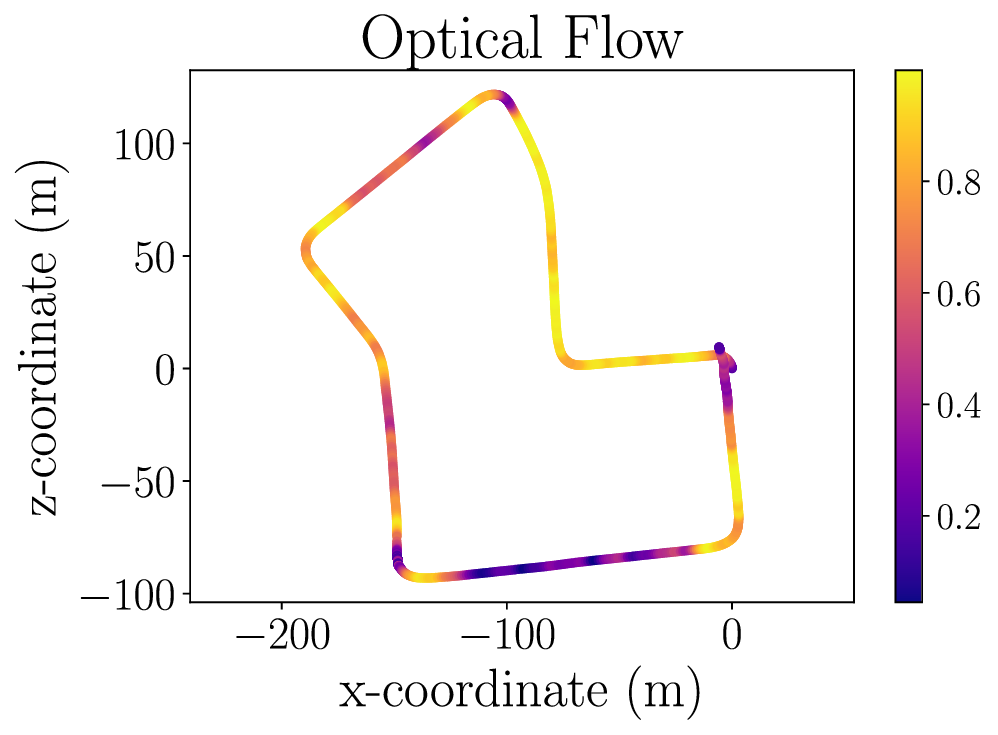}
        \caption{Optical Flow}
        \label{fig:07_flow}
    \end{subfigure}
    \begin{subfigure}[t]{0.225\textwidth}
        \centering
        \includegraphics[width=\textwidth]{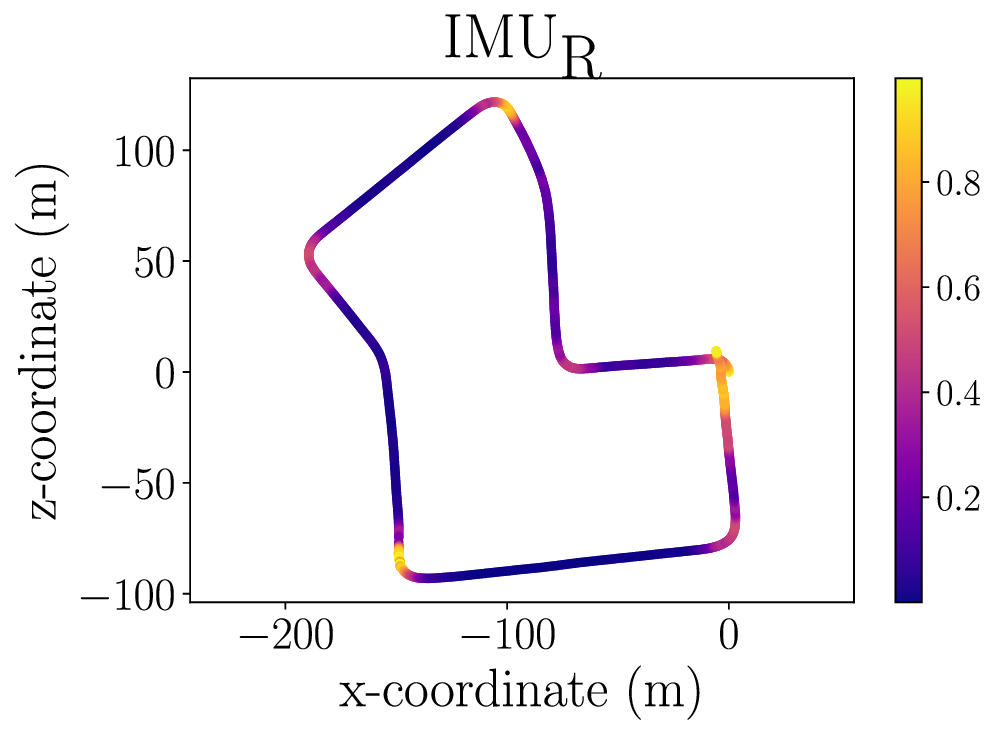}
        \caption{IMU Rotation}
        \label{fig:07_imur}
    \end{subfigure}
    
    \begin{subfigure}[t]{0.225\textwidth}
        \centering
        \includegraphics[width=\textwidth]{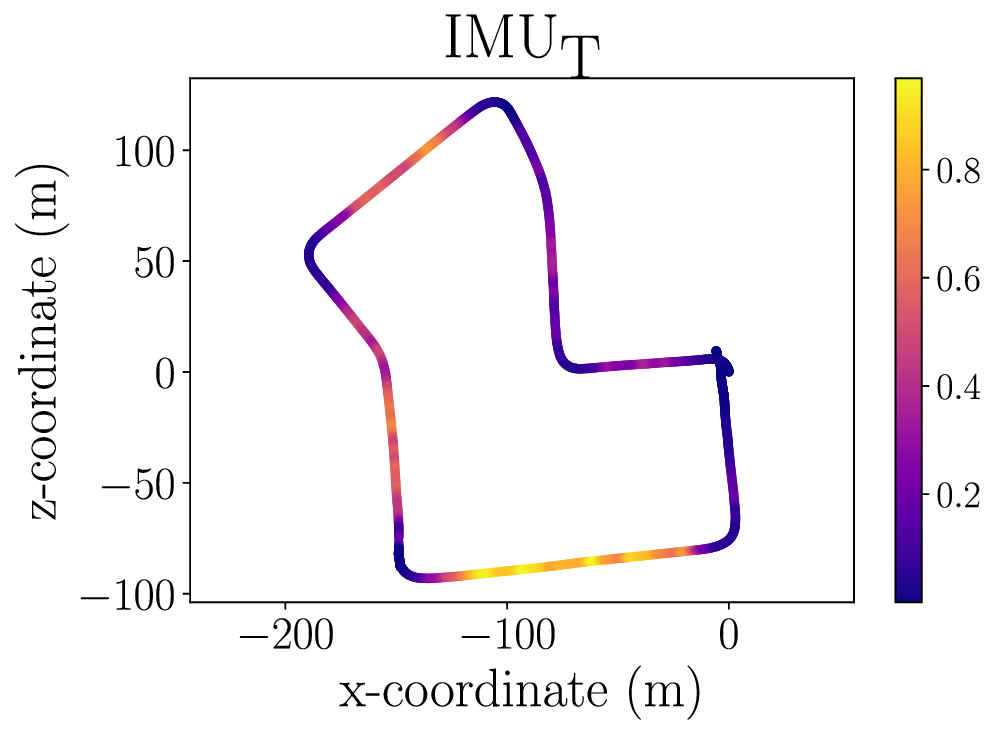}
        \caption{IMU Translation}
        \label{fig:07_imut}
    \end{subfigure}
    \begin{subfigure}[t]{0.225\textwidth}
        \centering
        \includegraphics[width=\textwidth]{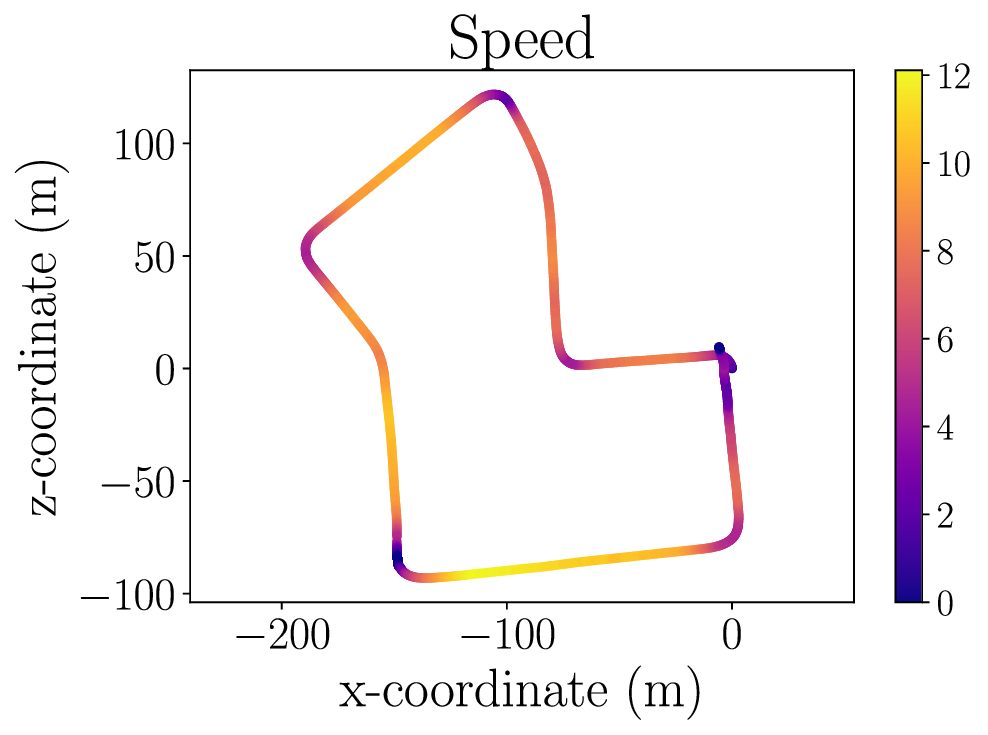}
        \caption{Speed (for reference, not modality)}
        \label{fig:07_speed}
    \end{subfigure}
    
    \caption{Self-emergent sensor weighting along Sequence 07 of KITTI Dataset \cite{kittiodometry}.}
    \label{fig:policy_07}
\end{figure}

One feature of XIRVIO is its ability to generate human-interpretable sensor weighting without any guidance from the loss functions. Figure \ref{fig:07_flow}, \ref{fig:07_imur}, and \ref{fig:07_imut} show the self-emergent weighting of different modalities along the path of KITTI Sequence 07 \cite{kittiodometry}, with Figure \ref{fig:07_speed} showing the speed along the path. It is observed that the model learns to prioritise optical flow in slow sections and not strictly connected to straight lines or corners. As for the IMU modalities, the model learns to favour rotation in low speeds or corners, while IMU translation is favoured more in sections with higher speeds and larger accelerations/decelerations. This is as expected as the IMU translation component in the KITTI dataset is measured in terms of acceleration, while the rotational component is in terms of angular velocity \cite{kittiodometry}.

While this paper uses a simpler set of modalities, this demonstrates XIRVIO's potential in helping users understand the model behaviour under different context, and how each sensory input contributes to the model's final pose estimation. It is highly beneficial to have an explainable machine learning framework in safety-critical applications.

\section{Conclusion}

The XIRVIO model is presented which is a transformer-based WGAN framework that features generative and iterative refinement of pose estimation, and adaptive sensor weighting for robust VIO applications. The generator encodes visual and IMU inputs, and iteratively produces refined "delta poses" that sum up to the final pose estimation. The critic evaluates intermediate outputs, enabling both WGAN-based training and self-assessment of pose quality. The experiments with the KITTI dataset confirm XIRVIO’s strong performance, matching state of the art learning-based methods. One of the highlights of XIRVIO is its Policy Encoder, which autonomously learns an adaptive weighting strategy for each sensor stream. This design not only improves pose accuracy but also provides explainability which is an essential aspect for safety-critical VIO applications. Although the current experiments focus on RGB images and a single inertial configuration, XIRVIO demonstrates the potential to provide explainable predictions under a more complex set of sensor modalities.

\bibliography{ref.bib}{}
\bibliographystyle{ieeetr}

\end{document}